\documentclass{article}

\usepackage{arxiv}
\usepackage[utf8]{inputenc} % allow utf-8 input
\usepackage[T1]{fontenc}    % use 8-bit T1 fonts
\usepackage{url}            % simple URL typesetting
\usepackage{booktabs}       % professional-quality tables
\usepackage{amsfonts}       % blackboard math symbols
\usepackage{nicefrac}       % compact symbols for 1/2, etc.
\usepackage{microtype}      % microtypography
\usepackage{lipsum}

\usepackage{subcaption}
\usepackage{graphicx}
\graphicspath{ {./images/} }
\usepackage{graphicx}
\usepackage{amsmath}
\usepackage{amssymb}
\usepackage{booktabs}
\usepackage{multirow}

\usepackage[linesnumbered,ruled,vlined]{algorithm2e}%[ruled,vlined]{
\usepackage{algpseudocode}

\usepackage[pagebackref,breaklinks,colorlinks]{hyperref}
\usepackage[capitalize]{cleveref}

\title{FedDM: Enhancing Communication Efficiency and Handling Data Heterogeneity in Federated Diffusion Models}

\author{
Jayneel Vora$^{1}$, Nader Bouacida$^{1}$,Aditya Krishnan$^{1}$,  Prasant Mohapatra$^{2}$\\
$^{1}$Department of Computer Science, University of California, Davis\\
$^{2}$Department of Computer Science, University of South Florida,Tampa\\
\texttt{$^{1}$\{jrvora,nbouacida,adikrishnan\}@ucdavis.edu}\\
\texttt{$^{2}$pmohapatra@usf.edu}
}

\begin{document}
\maketitle

\begin{abstract}
We introduce FedDM, a novel training framework designed for the federated training of diffusion models. Our theoretical analysis establishes the convergence of diffusion models when trained in a federated setting, presenting the specific conditions under which this convergence is guaranteed. We propose a suite of training algorithms that leverage the U-Net architecture as the backbone for our diffusion models. These include a basic Federated Averaging variant, FedDM-vanilla, FedDM-prox to handle data heterogeneity among clients, and FedDM-quant, which incorporates a quantization module to reduce the model update size, thereby enhancing communication efficiency across the federated network.

We evaluate our algorithms on FashionMNIST (28x28 resolution), CIFAR-10 (32x32 resolution), and CelebA (64x64 resolution) for DDPMs, as well as LSUN Church Outdoors (256x256 resolution) for LDMs, focusing exclusively on the imaging modality. Our evaluation results demonstrate that FedDM algorithms maintain high generation quality across image resolutions. At the same time, the use of quantized updates and proximal terms in the local training objective significantly enhances communication efficiency (up to 4x) and model convergence, particularly in non-IID data settings, at the cost of increased FID scores (up to 1.75x).
\end{abstract}

\keywords{federated \and diffusion \and \and communication efficiency \and quantized \and heterogeneity \and  skewness}

\section{Introduction}
Diffusion models\cite{song2019generative} represent an emerging class of generative neural networks that produce data from a training distribution through an iterative denoising process. Initially formalized as diffusion probabilistic models\cite{sohl2015deep}, their development advanced significantly with the introduction of denoising probabilistic models\cite{ho2020denoising}. Subsequent work demonstrated that diffusion models could incorporate conditionals to guide the synthesis process\cite{nichol2021improved}. The denoising diffusion implicit model (DDIM) introduced a non-Markovian diffusion process, achieving the same training objective with faster sampling\cite{song2020denoising}. Additionally, operating the diffusion process in latent space instead of pixel space has further enhanced their efficiency and effectiveness by ensuring reduced dimensionality of data\cite{rombach2022high}.

Diffusion models have rapidly become the de facto method for generating high-resolution images and have been extended to various tasks. These include controllable image generation\cite{kwon2022diffusion}, image segmentation\cite{baranchuk2021label}, image inpainting\cite{corneanu2024latentpaint}, image super-resolution\cite{yue2024resshift}, image restoration\cite{xia2023diffir}, anomaly detection\cite{wolleb2022diffusion}, and image-to-image translation\cite{saharia2022palette}. Their applications extend into bioinformatics for single-cell image analysis\cite{guo2024diffusion} and small molecular generation\cite{huang2023mdm}. Beyond image generation, diffusion models are also being utilized for text\cite{chen2024textdiffuser}, audio\cite{huang2023make}, and video\cite{ho2022video} modalities, demonstrating their versatility and broad applicability in various domains.

Diffusion models demonstrate several significant advantages over prior approaches such as Generative Adversarial Networks (GANs) \cite{goodfellow2020generative} and Variational Autoencoders (VAEs) \cite{kingma2013auto}. For one, diffusion models are more accessible to scale and condition, facilitating their adaptation to various data modalities and tasks\cite{zhang2023adding}. Unlike GANs, which suffer from model collapse due to their adversarial training dynamics, diffusion models avoid this issue entirely, leading to more stable and reliable training processes. Additionally, the iterative nature of diffusion models enhances their interpretability, providing a clearer understanding of the generative process\cite{nichol2021glide}. This iterative refinement allows diffusion models to produce higher-quality and more diverse samples\cite{ramesh2022hierarchical} compared to VAEs, which often face a trade-off between reconstruction quality and regularization\cite{lin2019balancing}.

\begin{figure}
    \centering
    \begin{subfigure}[b]{0.45\textwidth}
        \centering
        \includegraphics[width=\textwidth]{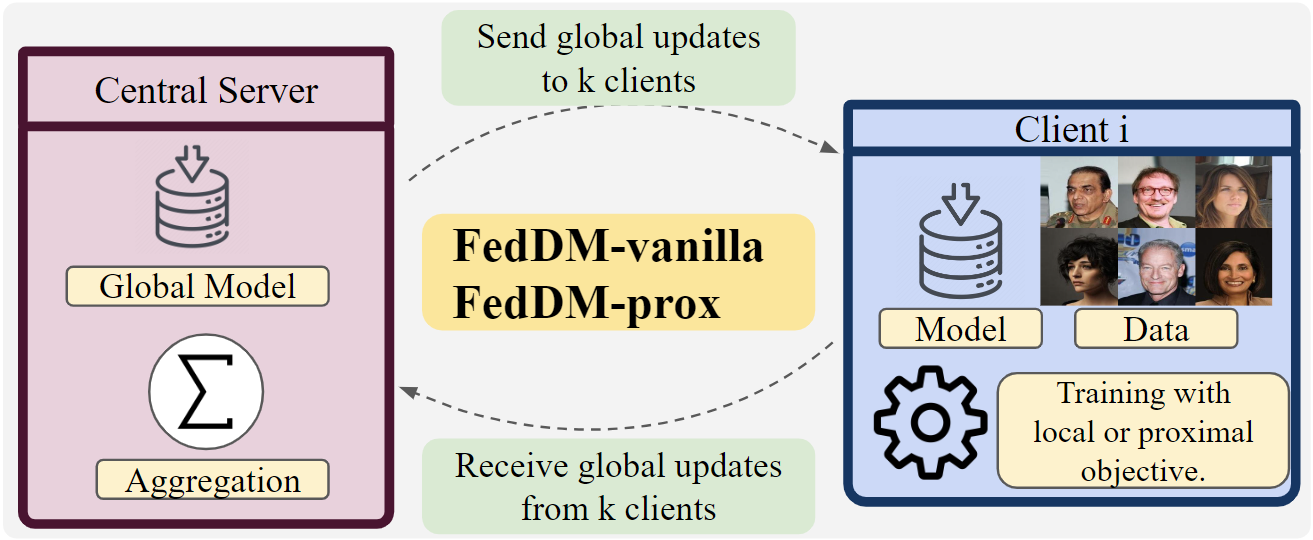}
        \caption{Fed-vanilla and Fed-prox}
        \label{fig:fedvanilla}
    \end{subfigure}
    \hspace{0.05\textwidth}
    \begin{subfigure}[b]{0.45\textwidth}
        \centering
        \includegraphics[width=\textwidth]{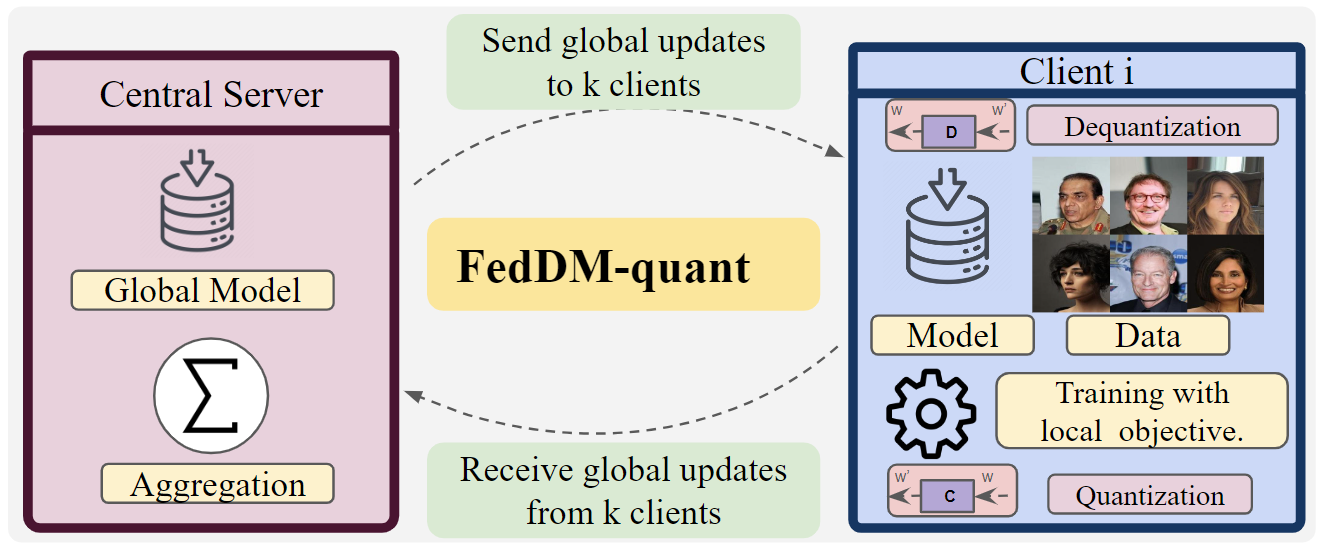}
        \caption{Fed-quant}
        \label{fig:fedquant}
    \end{subfigure}
    \caption{Simplified visualization of FedDM training algorithms presented in this work.}
    \label{fig:fedvariants}
\end{figure}
Even with this progress, diffusion models necessitate substantial data for effective training\cite{wang2024patch}, yet most organizations lack such extensive datasets\cite{kumari2023ablating} and end up using copyrighted material\cite{chesterman2024good}. Organizations seek to train models using aggregated data collaboratively while avoiding direct data sharing. Data is often dispersed across various sources, with each source possessing datasets that are insufficient in size and diversity to locally train a high-quality diffusion model reflective of the entire population of distributed data. Moreover, privacy constraints often deny the possibility of central pooling or sharing of data\cite{mattioli2017data}. 

Federated learning (FL)\cite{mcmahan2017communication} is a distributed machine learning approach where multiple devices train a shared model collaboratively while keeping their local data decentralized. It aims to minimize a global objective by combining local updates from each device, thus enhancing privacy and reducing communication overhead. Consequently, the development of a federated diffusion model algorithm is imperative. Such an algorithm would enable training a diffusion model that accurately represents the entire distributed data while preserving data privacy. 
% Federated diffusion models also facilitate the generation of publicly accessible data, thereby leveraging the collective data without compromising privacy.

Despite early research demonstrating that diffusion models can be trained in a federated manner, significant gaps remain in the literature. The existing body of work is limited; thus far, experiments have primarily utilized low-resolution datasets, failing to address the challenges associated with high-resolution data\cite{de2024training}. Additionally, these studies are limited to Denoising Diffusion Probabilistic Models (DDPMs)\cite{stanley2024phoenix}, neglecting other types of diffusion models such as DDIMs or Latent Diffusion Models (LDMs). There has been no investigation into the privacy implications of federated diffusion models, a critical aspect given the sensitive nature of many datasets\cite{carlini2023extracting}. Furthermore, developing communication-efficient variants of federated learning for diffusion models remains unexplored, an essential factor for practical deployment.

To solve this challenge, we propose federated training algorithms as depicted in Figure \ref{fig:fedvariants}: (a)FedDM-vanilla, FedDM-prox, and (b) FedDM-quant and explore various implications affecting the convergence of diffusion models.
Our research contributions can be summarized as follows:
\begin{itemize}
    \item We introduce "Fed-DM," a novel training framework that enables effective federated learning for diffusion models using the U-Net backbone architecture.
    \item We provide theoretical proof demonstrating that diffusion models can achieve convergence under federated training conditions, validating the feasibility and stability of applying federated learning to these models.
    \item We enhance communication efficiency in federated networks by incorporating post-training quantized model updates, showing that this approach results in minimal degradation in generation quality.
    \item Our comprehensive evaluations across diverse datasets and diffusion models, exploring a wide range of federated learning hyperparameters, validate the effectiveness and robustness of our proposed framework.
\end{itemize}

\section{Related Work}

\subsection{Communication Efficiency in FL}
FL enhances data privacy and security by training models in a decentralized manner, but frequent model updates between clients and the server create significant communication overhead, making communication efficiency a crucial research focus for scalability and cost-effectiveness. 
Model compression techniques such as quantization, sparsification, and knowledge distillation are prominent approaches to improve communication efficiency in Federated Learning (FL). FedComLoc \cite{yi2024fedcomloc} improves FL communication efficiency by integrating compression techniques like Top-K and quantization into local training, reducing overheads in heterogeneous settings. FediAC\cite{su2024expediting} utilizes client voting and model aggregation to improve communication efficiency in FL. FedSZ\cite{wilkins2023efficient} is a lossy compression-based framework, incorporating techniques like data partitioning, lossy model parameters, and efficient serialization to enhance communication efficiency. Top-k SGD \cite{lu2023top} transmits only the most significant gradient updates, effectively reducing the data volume exchanged. Federated distillation methods, such as FedMD \cite{li2019fedmd}, minimize the need to transmit full model parameters by transferring knowledge from a large global model to smaller local models using a public dataset.
Additionally, optimizing communication rounds has been explored in methods like FedAvgM \cite{hsu2019measuring}, which introduces momentum updates to stabilize and accelerate convergence, thereby reducing the number of necessary communication rounds. Client selection techniques, such as FedCS \cite{nishio2019client}, enhance overall efficiency by dynamically choosing clients with higher computational capabilities or better network conditions. Hierarchical FL structures, as proposed by HierFAVG \cite{liu2020client}, aggregate models at intermediary edge servers before sending them to the central server, thus decreasing the communication burden on the core network.

\subsection{Quantized FL Model Updates}
Quantization is another practical approach to reducing the communication load by lowering the precision of model parameters, for example, from int32 to int16 or int8. Alistarh et al. \cite{alistarh2017qsgd} introduces a quantization scheme that adjusts the quantization level according to gradient sparsity. Jhunjhunwala et al. \cite{jhunjhunwala2021adaptive} implement an adaptive quantization strategy, AdaQuantFL, that varies the quantization levels throughout the training process. Huang et al. \cite{huang2023wireless} demonstrate that quantizing model updates in asynchronous federated learning yields a provable upper bound on quantization error and the number of bits required for transmission. Reisizadeh et al. \cite{reisizadeh2020fedpaq} develop FedPAQ, applying QSGD to local updates before transmission, balancing communication reduction with acceptable performance degradation. Chen et al. \cite{chen2024mixed} propose FedMPQ, a mixed precision strategy for local updates, while Liu et al. \cite{liu2022hierarchical} establish tighter bounds for hierarchical FL algorithms. Chen et al. \cite{chen2021dynamic} introduce FEDHQ and its dynamic enhancement FEDHQ+, which optimally aggregate updates from clients with varying quantization precision levels by assigning aggregation weights inversely proportional to the normalized quantization error. Honig et al. \cite{honig2022dadaquant} leverage time- and client-adaptive quantization to enhance communication efficiency in FL through sophisticated integration of QSGD-based stochastic fixed-point quantization. Bouzinis et al. \cite{bouzinis2023wireless} rigorously analyze FL with stochastic quantization of local model parameters, offering insights into the optimality gap, and present a convex optimization approach that jointly optimizes computation and communication resources to minimize convergence time under energy consumption and quantization error constraints. Shlezinger et al. \cite{shlezinger2020uveqfed} utilize subtractive dithered lattice quantization \cite{zamir1992universal} for universal model update encoding, enabling accurate server-side recovery of updates transmitted over rate-constrained channels.

\subsection{Federated Diffusion Models}
Jothiraj et al. \cite{stanley2024phoenix} introduced Phoenix, the pioneering work on federated learning with diffusion models (FL-DM), which aggregates models at a central server for Denoising Diffusion Probabilistic Models (DDPMs), assuming universal client participation. Li et al. \cite{li2024feddiff} proposed FedDiff for federated multi-modal remote sensing land cover classification. FedDiff employs a dual-branch diffusion process, multi-head self-attention, and frequency domain parsing within a U-Net architecture, leveraging low-rank singular value decomposition (SVD) for feature map compression and gradient averaging to enhance performance and reduce communication costs. Allmendinger et al. \cite{allmendinger2024collafuse} presented CollaFuse, a split learning framework executing computationally intensive steps on a shared server. de Goede et al. \cite{de2024training} introduced FedDiffuse, which explores training algorithms based on U-Net architecture that share different parts of the U-Net and employ FedAvg for model averaging, evaluated on 28x28 black-and-white images. Both Phoenix and CollaFuse were evaluated on low-resolution image datasets based on DDPMs. In contrast, we evaluate Latent Diffusion Models (LDMs) on a high-resolution dataset.

\section{Method}
% \section{Method}
\subsection{Preliminaries}
\subsubsection{Diffusion Models}
Diffusion models are generative models that learn data distributions by reversing a stochastic process that gradually perturbs the data \cite{sohl2015deep}. The forward process is a Markov chain adding Gaussian noise to the data, defined as:
\begin{equation}
    q(\mathbf{x}_{t}|\mathbf{x}_{t-1}) = \mathcal{N}(\mathbf{x}_{t}; \sqrt{1-\beta_t} \mathbf{x}_{t-1}, \beta_t \mathbf{I}),
\end{equation}
Where $\beta_t$ is the variance schedule. The overall process is:
\begin{equation}
    q(\mathbf{x}_{1:T}|\mathbf{x}_0) = \prod_{t=1}^T q(\mathbf{x}_t|\mathbf{x}_{t-1}).
\end{equation}
The reverse process, parameterized by a neural network $\mathbf{\mu}_\theta(\mathbf{x}_t, t)$, predicts the mean of the reverse transition:
\begin{equation}
    p_\theta(\mathbf{x}_{t-1}|\mathbf{x}_{t}) = \mathcal{N}(\mathbf{x}_{t-1}; \mathbf{\mu}_\theta(\mathbf{x}_t, t), \sigma_t \mathbf{I}).
\end{equation}
Training involves maximizing the variational lower bound (VLB) on the data likelihood:
\begin{equation}
    \mathbb{E}_{q(\mathbf{x}_{0:T})} \left[ \log p_\theta(\mathbf{x}_0) - \sum_{t=1}^T \text{KL}\left( q(\mathbf{x}_{t-1}|\mathbf{x}_t, \mathbf{x}_0) || p_\theta(\mathbf{x}_{t-1}|\mathbf{x}_t) \right) \right].
\end{equation}

Sampling involves reversing the forward process, starting from a random noise sample $\mathbf{x}_T \sim \mathcal{N}(0, \mathbf{I})$ and iteratively denoising:
\begin{equation}
    \mathbf{x}_{t-1} = \mathbf{\mu}_\theta(\mathbf{x}_t, t) + \sigma_t \mathbf{z},
\end{equation}
where $\mathbf{z} \sim \mathcal{N}(0, \mathbf{I})$. This process is repeated until $\mathbf{x}_0$ is obtained.

Latent Diffusion Models (LDMs) extend diffusion models by operating in a latent space rather than the pixel space \cite{rombach2022high}. The high-dimensional data $\mathbf{x}_0$ is encoded into a lower-dimensional latent representation $\mathbf{z}_0 = \mathcal{E}(\mathbf{x}_0)$ using an encoder $\mathcal{E}$. The diffusion process in latent space is:
\begin{equation}
    q(\mathbf{z}_{t}|\mathbf{z}_{t-1}) = \mathcal{N}(\mathbf{z}_{t}; \sqrt{1-\beta_t} \mathbf{z}_{t-1}, \beta_t \mathbf{I}),
\end{equation}
With the reverse process:
\begin{equation}
    p_\theta(\mathbf{z}_{t-1}|\mathbf{z}_{t}) = \mathcal{N}(\mathbf{z}_{t-1}; \mathbf{\mu}_\theta(\mathbf{z}_t, t), \sigma_t \mathbf{I}).
\end{equation}
After the reverse process, the latent representation $\mathbf{z}_0$ is decoded back to the pixel space using a decoder $\mathcal{D}$: $\mathbf{x}_0 = \mathcal{D}(\mathbf{z}_0)$. This approach benefits from operating in a lower-dimensional space, improving efficiency and scalability.

\subsubsection{Federated Averaging}
Federated Averaging (FedAvg) \cite{mcmahan2017communication} allows decentralized model training across clients without sharing local data. Consider a global model $\mathbf{w}$ being trained across $K$ clients. Each client $i$ has dataset $\mathcal{D}_i$ and performs local updates on the global update they receive$\mathbf {w}_i$. In each round $t$, a subset of $k$ clients is selected for aggregation($\text{where } 1\leq k \leq K$. The global model update is $\mathbf{w}^{(t)} = \frac{1}{k} \sum_{i \in S_t} n_i \mathbf{w}_i^{(t)}$, where $n_i = \frac{|\mathcal{D}_i|}{\sum_{j \in S_t} |\mathcal{D}_j|}$ and $S_t$ is the set of selected clients where $k=|S_t|$. Each client $i$ performs local optimization: $\mathbf{w}_i^{(t+1)} = \mathbf{w}_i^{(t)} - \eta \nabla \ell_i(\mathbf{w}_i^{(t)}, \mathcal{D}_i)$, where $\eta$ is the learning rate and $\ell_i$ is the local loss function. The global model is updated by weight-averaging the local updates.

\subsubsection{Post-Training Quantization}
Post-training quantization (PTQ) compresses neural networks by reducing the precision of weights and activations. Let $\mathbf{W} \in \mathbb{R}^{n \times m}$ be the weight matrix. PTQ maps $\mathbf{W}$ to lower precision $\hat{\mathbf{W}} \in \mathbb{Q}^{n \times m}$, where $\mathbb{Q}$ is the quantized space: $\hat{\mathbf{W}} = \text{round}( \frac{\mathbf{W} - \min(\mathbf{W})}{\Delta} ) \Delta + \min(\mathbf{W})$, with $\Delta = \frac{\max(\mathbf{W}) - \min(\mathbf{W})}{2^b - 1}$, where $b$ is the number of bits. The objective is to minimize quantization error: $\mathcal{L}_{\text{quant}} = \|\mathbf{W} - \hat{\mathbf{W}}\|_2^2$. PTQ of model updates in federated learning enhances communication efficiency by reducing data transmission between clients and the central server, making federated models practical.

\subsection{Approach 1: FedDM-vanilla}
\subsubsection{Problem Formulation}
We consider a federated learning scenario involving $K$ clients, each possessing a local dataset $\mathcal{D}_k$ for $k \in \{1, \ldots, K\}$. The objective is to collaboratively train a global denoising probabilistic diffusion model while preserving the privacy of local datasets. 
\begin{algorithm}
  \caption{Algorithm 1: FedDM-vanilla}
  \label{algo:algo1}
  \footnotesize
  \KwIn{Initial model parameters \( \theta^0 \), number of total clients \( K \), number of selected clients \( k \), local epochs \( E \), learning rate \( \eta \), global epochs \( R \)}
  \# Initialize global model parameters\\
  Initialize \( \theta^0 \)\\
  \For{each round \( r = 0, 1, \ldots, R-1 \)}
  {
    \# Server distributes \( \theta^r \) to selected clients\\
    Randomly select \( k \) clients from \( K \) clients\\
    \For{\textbf{each client} \( j \in \{ 1, \ldots, k \} \) \textbf{in parallel}}
    {
        Initialize \( \theta_j^{r+1} \) with \( \theta^r \)\\
        \For{epoch \( e = 1, \ldots, E \)}
        {
            \# Perform local update\\
            \( \theta_j^{r+1} \leftarrow \theta_j^{r+1} - \eta \nabla \mathcal{L}_j(\theta_j^{r+1}) \)\\
        }
    }
    \# Server aggregates the updated parameters\\
    \( \theta^{r+1} \leftarrow \frac{1}{k} \sum_{j=1}^{k} n_j\theta_j^{r+1} \)\\
  }
  \textbf{return} Global model parameters \( \theta^R \)
\end{algorithm}
Let $x \sim q(x)$ denote the data distribution, and let $p_\theta(x_t | x_{t-1})$ represent the diffusion process, where $\theta$ denotes the model parameters. The diffusion model aims to learn a denoising function $\epsilon_\theta(x_t, t)$, such that the reverse process $p_\theta(x_{t-1} | x_t)$ approximates $q(x_{t-1} | x_t)$. The optimization objective involves minimizing the negative log-likelihood of the data, defined as:

\[
\mathcal{L}(\theta) = \mathbb{E}_{q(x_0)} \left[ \sum_{t=1}^{T} \mathbb{E}_{q(x_t | x_{t-1})} \left[ -\log p_\theta(x_{t-1} | x_t) \right] \right].
\]

\subsubsection{Federated Aggregation Algorithm}
In a federated setting, each client $k$ locally optimizes the following loss function:

\[
\mathcal{L}_k(\theta) = \mathbb{E}_{q_k(x_0)} \left[ \sum_{t=1}^{T} \mathbb{E}_{q_k(x_t | x_{t-1})} \left[ -\log p_\theta(x_{t-1} | x_t) \right] \right],
\]

where $q_k(x)$ represents the data distribution of client $k$. The central server then aggregates the local updates to form a global model.

The federated averaging algorithm is adapted to train diffusion models by iteratively aggregating local model updates as described in Algorithm \ref{algo:algo1}. 
Initially, the global model parameters $\theta^0$ are initialized and distributed to all $K$ clients. During each global iteration $r$, a subset $k$ of the clients performs local updates based on their respective datasets $\mathcal{D}_k$. 
Specifically, each client $k$ receives the current global model parameters $\theta^r$ and performs $E$ epochs of local training, updating the parameters according to the gradient of the local loss function:
\[
\theta_k^{r+1} \leftarrow \theta^r - \eta \nabla \mathcal{L}_k(\theta^r),
\]
Where $\eta$ denotes the learning rate, after completing the local updates, the central server aggregates the updated parameters from the selected $k$ clients to form the new global parameters:
\[
\theta^{r+1} \leftarrow \frac{1}{k} \sum_{j=1}^{k} n_j\theta_j^{r+1}.
\]
% The federated training process is summarized in 

\subsubsection{Convergence Analysis}

Let \( \{ x_t \} \) denote the sequence of denoised images at iteration \( t \), and let \( \epsilon \) be the denoising function. We assume that \( \epsilon \) is Lipschitz continuous with constant \( L < 1 \), i.e., \( \| \epsilon(x) - \epsilon(y) \| \leq L \| x - y \| \) for all \( x, y \in \mathbb{R}^d \). The noise \( \zeta_t \) at iteration \( t \) has zero mean and bounded variance, i.e., \( \mathbb{E}[\zeta_t] = 0 \) and \( \mathbb{E}[\|\zeta_t\|^2] \leq \sigma^2 \). At each iteration \( t \), only \( k \) out of \( K \) clients are used to compute the global model update, given by \(\Delta x_t = \frac{1}{k} \sum_{i=1}^{k} \epsilon_i(x_t) + \zeta_t \), where \( \epsilon_i \) represents the local update for the denoising function of the \( i \)-th client.

We first show that the global denoising function \( \epsilon \) remains a contraction mapping under the federated averaging scheme. Let \( \epsilon_i \) be Lipschitz continuous with constant \( L_i < 1 \) for each client \( i \). The aggregated global update function \(\bar{\epsilon}(x) = \frac{1}{k} \sum_{i=1}^{k} n_i\epsilon_i(x) \) is also Lipschitz continuous with constant \( \bar{L} < 1 \). For any \( x, y \in \mathbb{R}^d \), we have \(\| \bar{\epsilon}(x) - \bar{\epsilon}(y) \| \leq \bar{L} \| x - y \| \), where \( \bar{L} = \frac{1}{k} \sum_{i=1}^{k} L_i < 1 \).

By the Banach Fixed-Point Theorem, \( \bar{\epsilon} \) has a unique fixed point \( x^* \) such that \( \bar{\epsilon}(x^*) = x^* \). Consider the iterative update rule \( x_{t+1} = \bar{\epsilon}(x_t) + \zeta_t \). Let \( x^* \) be the fixed point of \( \bar{\epsilon} \). Then, the distance from \( x_{t+1} \) to \( x^* \) can be bounded by \( \bar{L} \) times the distance from \( x_t \) to \( x^* \) plus the noise term \( \zeta_t \).

Taking expectations and noting that \( \mathbb{E}[\zeta_t] = 0 \), we get an expected contraction property. Over iterations, this leads to the bound \( \mathbb{E}[\| x_t - x^* \|] \leq \bar{L}^t \mathbb{E}[\| x_0 - x^* \|] + \frac{\sigma}{1 - \bar{L}} \). As \( t \to \infty \), \( \bar{L}^t \) tends to zero, implying that \( \mathbb{E}[\| x_t - x^* \|] \) converges to zero. Therefore, the sequence \( \{ x_t \} \) converges to the fixed point \( x^* \) in expectation.

\subsection{Approach 2:FedDM-prox}
Independent and Identically Distributed(IID) data refers to data samples independently drawn from the same underlying distribution, ensuring uniform representation across the dataset. 
In contrast, non-IID data consists of samples that are not independent and exhibit varying distributions, contributing to data heterogeneity. 
Label-skewed data exacerbates the challenges of federated learning algorithms by introducing significant data heterogeneity, which can lead to biased model updates and reduced generalizability across diverse client data distributions\cite{lu2024federated}

To address the challenges posed by non-IID data distributions across clients, we introduce a proximal term to the local objective function inspired by the FedProx algorithm\cite{li2018convergence}. This term penalizes deviations from the global model parameters, thus maintaining convergence and stability. Specifically, the local objective function for each client is modified by adding a proximal term: 

\begin{align*}
\mathcal{L}_k^{\text{prox}}(\theta) &= \mathbb{E}_{q_k(x_0)} \left[ \sum_{t=1}^{T} \mathbb{E}_{q_k(x_t | x_{t-1})} \left[ -\log p_\theta(x_{t-1} | x_t) \right] \right] \\
&\quad + \frac{\mu}{2} \|\theta - \theta^r\|^2,
\end{align*}

where \( \mu \) is the proximal term coefficient, \( \theta \) are the local model parameters, and \( \theta^r \) are the global model parameters from the previous round. During local updates, the parameters are updated according to:

\[
\theta_j^{r+1} \leftarrow \theta_j^{r+1} - \eta \left( \nabla \mathcal{L}_j(\theta_j^{r+1}) + \mu (\theta_j^{r+1} - \theta^r) \right),
\]

where \( \eta \) denotes the learning rate and \( \mu \) is the proximal term coefficient. The rest of the federated aggregation process remains as described in Algorithm \ref{algo:algo1}.

By incorporating the proximal term in the local training objective, FedDM-Prox mitigates the impact of data heterogeneity among clients. The proximal term helps to stabilize the training process by constraining the local updates to remain close to the current global model parameters. 

\subsection{Approach 3: FedDM-Quant}
To enhance communication efficiency in federated learning, we propose a quantized update mechanism, FedDM-Quant, which transmits quantized model updates between clients and the central server. This approach involves each client dequantizing the received global model, training it locally with full precision for several epochs, calibrating the quantization parameters, and then re-quantizing the updated model before sending it back to the central server.

\begin{algorithm}
\caption{FedDM-Quant}
\label{algo
}
\footnotesize
\KwIn{Initial model parameters $\theta^0$, number of clients $K$, number of contributing clients $k$, global epochs $R$, local epochs $E$, learning rate $\alpha$, quantization function $Q(\cdot)$, dequantization function $D(\cdot)$, calibration function $C(\cdot)$}
Initialize $\theta^0$\
\For{each round $r = 0, 1, \ldots, R-1$}
{
$\hat{\theta}^r \leftarrow Q(\theta^r)$\
Select $k$ clients from $K$ clients\
\For{\textbf{each client} $i \in { 1, \ldots, k }$ \textbf{in parallel}}
{
$\theta_i^{r+1} \leftarrow D(\hat{\theta}^r)$\
\For{epoch $e = 1, \ldots, E$}
{
$\theta_i^{r+1} \leftarrow \theta_i^{r+1} - \alpha \nabla \mathcal{L}_i(\theta_i^{r+1})$\
}
$\theta_i^{r+1} \leftarrow C(\theta_i^{r+1})$\\
Sample $N$ images uniformly through the sampling process\ \\
Adjust quantization scales and zero points to minimize quantization error based on sampled images\\
$\hat{\theta}i^{r+1} \leftarrow Q(\theta_i^{r+1})$\
}
$\hat{\theta}^{r+1} \leftarrow \frac{1}{k} \sum{i=1}^{k} n_i\hat{\theta}_i^{r+1}$\
$\theta^{r+1} \leftarrow D(\hat{\theta}^{r+1})$\
}
\textbf{return} Global model parameters $\theta^R$
\end{algorithm}

\begin{table*}[ht]
\caption{
% A comparison of the best quality of generations across various total and contributing clients for the best settings for global and local epochs for Fed-vanilla.
Comparison of the best generation quality across varying total and contributing clients for optimal global and local epoch settings using Fed-vanilla on CIFAR-10 and LSUN-Church datasets.
}
\label{tab:Fed-vanilla}
\centering
\scalebox{1.0}
{
\begin{tabular}{l|c|c|c|c|c|c}
\hline 
\multirow{2}{*}{\textbf{Dataset}} & \multirow{2}{*}{\textbf{Model}} & \multicolumn{2}{c|}{\textbf{Clients}} & \multirow{2}{*}{\textbf{\begin{tabular}[c]{@{}c@{}}Global\\ Epochs(R)\end{tabular}}} & \multirow{2}{*}{\textbf{\begin{tabular}[c]{@{}c@{}}Local\\ Epochs(E)\end{tabular}}} & \multirow{2}{*}{\textbf{\begin{tabular}[c]{@{}c@{}}FID\\ $(\downarrow)$\end{tabular}}} \\
\cline{3-4}
 &  & \textbf{\begin{tabular}[c]{@{}c@{}}Total\\ Clients\end{tabular}} & \textbf{\begin{tabular}[c]{@{}c@{}}Contributing\\ Clients\end{tabular}} &  &  &  \\
\hline
\multirow{5}{*}{CIFAR10} & \multicolumn{5}{c|}{DDPM trained centrally for 800 epochs} & 5.38 \\
\cline{2-7} 
 & \multirow{4}{*}{DDPM with Fed-vanilla} & \multirow{2}{*}{10} & 4 & 16 & 10 & 7.24 \\
\cline{4-7} 
 &  &  & 6 & 16 & 10 & 7.24 \\
\cline{3-7} 
 &  & \multirow{2}{*}{15} & 6 & 8 & 15 & \textbf{6.31} \\
\cline{4-7} 
 &  &  & 9 & 16 & 15 & 6.87 \\
\hline
\multirow{5}{*}{LSUN-Church} & \multicolumn{5}{c|}{DDPM trained centrally for 800 epochs} & 9.07 \\
\cline{2-7} 
 & \multirow{4}{*}{LDM with Fed-vanilla} & \multirow{2}{*}{10} & 4 & 8 & 15 & 13.39 \\
\cline{4-7} 
 &  &  & 6 & 4 & 10 & 12.49 \\
\cline{3-7} 
 &  & \multirow{2}{*}{15} & 6 & 16 & 10 & 14.33 \\
\cline{4-7} 
 &  &  & 9 & 8 & 10 & \textbf{11.79} \\
\hline
\end{tabular}
}
\end{table*}

\begin{figure}
    \centering
    \includegraphics[width=0.75\textwidth]{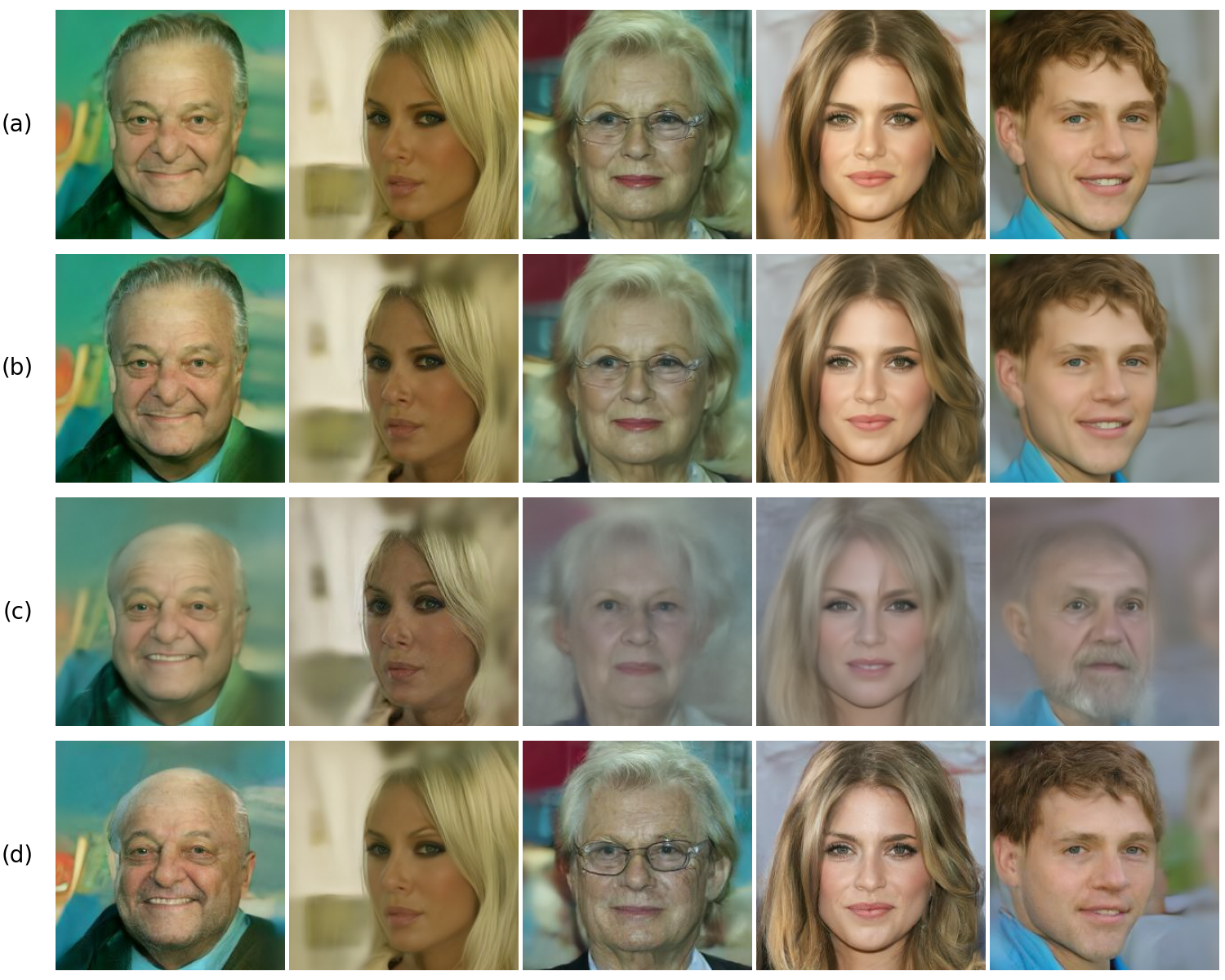}
    \caption{
    64x64 images sampled using the initial random noise using models trained on the CelebA dataset with the following variants: 
    (a) centralized 
    (b) FedDM-vanilla with 10 clients, 6 contributing clients, 20 global epochs and 15 local epochs.
    (c)  FedDM-vanilla with weights quantized to 8 bits, 10 clients, 6 contributing clients, 20 global epochs and 15 local epochs. 
    (d) FedDM-quant with weights quantized to 8 bits, 10 clients, 6 contributing clients, 20 global epochs and 15 local epochs. 
    }
    \label{fig:federated_averaging_variants}
\end{figure}
Each client begins with the global model $\hat{\theta}^r$, dequantizes it to $\theta_i^{r+1}$ using a dequantization function $D(\cdot)$. The client then performs local training, updating the model parameters $\theta_i^{r+1}$ over $E$ local epochs. After local training, the client calibrates the weight quantization parameters by sampling images uniformly. This post-training quantization (PTQ) algorithm is adapted from PTQ4DM \cite{shang2023post}, which is designed to minimize quantization error in image diffusion models. The calibrated and re-quantized model is then returned to the central server, completing the communication-efficient update cycle.
\section{Evaluations}
\subsection{Evaluation Setup}

\textbf{Datasets: }We employ a diverse set of datasets to evaluate our algorithms. We utilize the Fashion MNIST dataset \cite{xiao2017fashion} comprising 60,000 28x28 grayscale images of fashion items, CIFAR10 \cite{krizhevsky2009learning}, containing 50,000 32x32 RGB images across ten classes, and  CelebA \cite{liu2015deep} center-cropped to 64x64 RGB, consisting of 200,000 celebrity face images. To evaluate models on high-resolution images, we use LSUN Church Outdoors \cite{yu2015lsun} with 120,000 256x256 RGB images of church outdoor scenes.
%table 2
\begin{table*}[]
\centering
\caption{FID scores across increasing image resolution datasets for 10 clients (6 contributing), reporting the best settings for global and local epochs. }
\label{tab:scaling}
\begin{tabular}{l|c|ccc|c}
\hline
\textbf{Dataset} &
  \textbf{Resolution} &
  \multicolumn{1}{c|}{\textbf{Model}} &
  \multicolumn{1}{c|}{\textbf{\begin{tabular}[c]{@{}c@{}}Global \\ Epochs(R)\end{tabular}}} &
  \textbf{\begin{tabular}[c]{@{}c@{}}Local \\ Epochs(E)\end{tabular}} &
      \textbf{FID$(\downarrow)$}\\ \hline
      
\multirow{2}{*}{Fashion MNIST} & \multirow{2}{*}{28x28}   & \multicolumn{3}{c|}{DDPM trained centrally  for 800 epochs}                             & 5.07   \\ \cline{3-6} 
 
                               &                          & \multicolumn{1}{c|}{DDPM-Fed-vanilla} & \multicolumn{1}{c|}{12}&5&5.44\\ \hline
\multirow{2}{*}{CIFAR10} & \multirow{2}{*}{32x32}   & \multicolumn{3}{c|}{DDPM-Centralized  for 800 epochs}                             &5.38 \\ \cline{3-6} 
                               % &                          & \multicolumn{1}{c|}{DDPM-Fed-vanilla}  & \multicolumn{1}{c|}{} &  &  &  \\ \cline{3-7} 
                               &                          & \multicolumn{1}{c|}{DDPM-Fed-vanilla} & \multicolumn{1}{c|}{16} &10  & 6.16  \\ \hline

\multirow{2}{*}{CelebA} & \multirow{2}{*}{64x64}   & \multicolumn{3}{c|}{DDPM trained centrally  for 800 epochs}                             & 6.73 \\ \cline{3-6} 
                               % &                          & \multicolumn{1}{c|}{DDPM-Fed-vanilla}  & \multicolumn{1}{c|}{} &  &  &  \\ \cline{3-7} 
                               &                          & \multicolumn{1}{c|}{DDPM-Fed-vanilla} & \multicolumn{1}{c|}{20}&10&6.94\\ \hline
                               
\multirow{2}{*}{LSUN Church}& \multirow{2}{*}{256x256}   & \multicolumn{3}{c|}{LDM trained centrally  for 800 epochs}                             & 9.07 \\ \cline{3-6} 
                               &                          & \multicolumn{1}{c|}{LDM-Fed-vanilla}  & \multicolumn{1}{c|}{8} &10  & 12.49   \\ \cline{1-6} 

\end{tabular}
\end{table*}
%table 3
\begin{table*}[]
\centering
\caption{Local model weight quantization comparison between the Fed-vanilla algorithm (without PTQ calibration) and with Fed-quant across 32, 16, and 8-bit configurations, with activations at full precision, using a federated learning setup of 10 clients (6 contributing), and performance metrics reported for DDPM on CIFAR-10 and LDM-8 on LSUN-Church.}
\label{tab:quant}
\begin{tabular}{c|cc|c|c|c|c}
\hline
\textbf{Dataset} &
  \multicolumn{1}{c|}{\textbf{Model}} &
  \textbf{\begin{tabular}[c]{@{}c@{}}
  Weight\\ Bitwidth\end{tabular}} &
  \textbf{\begin{tabular}[c]{@{}c@{}}Global \\ Epochs(R)\end{tabular}} &
  \textbf{\begin{tabular}[c]{@{}c@{}}Local \\ Epochs(E)\end{tabular}} &
  \textbf{\begin{tabular}[c]{@{}c@{}}FID \\ $(\downarrow)$\end{tabular}} &
  \textbf{\begin{tabular}[c]{@{}c@{}}\ Mebibytes \\  transferred\end{tabular}} \\ \hline
\multirow{5}{*}{CIFAR10} & \multicolumn{1}{c|}{\multirow{2}{*}{DDPM-Fed-vanilla}}  & 32 &16  &10 &6.16 & 6549.99 \\ 
\cline{3-7} 
                          & \multicolumn{1}{c|}{}                                & 16 & 30 & 15 & 10.24 &   6140.61\\ 
                          
                          \cline{2-7} 
                          % & \multicolumn{1}{c|}{}                               & W8/A32  &  &  &  &  &  &  \\ \cline{2-9} 
                          & \multicolumn{1}{c|}{\multirow{2}{*}{DDPM-Fed-quant}} & 16& 16 &10  & \textbf{6.94} &  6549.99   \\ \cline{3-7} 
                          & \multicolumn{1}{c|}{}                               & 8  & 16 & 15 &  7.06&   \textbf{3274.99} \\ \hline

\multirow{5}{*}{LSUN-Church} & \multicolumn{1}{c|}{\multirow{2}{*}{LDM-Fed-vanilla}}  & 32 &16  &10  &9.07   &4733.86 \\ \cline{3-7} 
                          & \multicolumn{1}{c|}{}                               & 16 & 20 &15  & 18.15   &2958.66 \\ 
                          
                          \cline{2-7} 
                          % & \multicolumn{1}{c|}{}                               & W8/A32  &  &  &  &  &  &  \\ \cline{2-9} 
                          & \multicolumn{1}{c|}{\multirow{2}{*}{LDM-Fed-quant}}  & 16 &16  &10  &14.27   &3698.32\\ \cline{3-7} 
                          & \multicolumn{1}{c|}{}                               & 8  &8&15&15.13  &  3698.32 \\ \hline
\end{tabular}
\end{table*}\\
\textbf{Models: }Our evaluation focuses on two types of diffusion models: Denoising Diffusion Probabilistic Models (DDPMs) and Latent Diffusion Models (LDMs). DDPMs are assessed using the Fashion MNIST, CIFAR10, and CelebA datasets. For high-resolution tasks, we utilize LDMs, specifically with latent space reductions of $-8$, evaluated on the LSUN Church Outdoor dataset. All models undergo training for 1000 timesteps with a linear beta scheduler ranging from 0.0001 to 0.02.

\textbf{Testbed: }Evaluations are conducted on a workstation equipped with three NVIDIA RTX A6000 GPUs (49,140 MiB each) and dual Intel Xeon Silver 4214R CPUs (48 cores, 96 threads, 2.40GHz base, 3.50GHz turbo). The system has 250 GiB RAM, 2 GiB swap, and runs Ubuntu 20.04.6 LTS (kernel 5.15.0-92-generic). The testbed employs Python 3.8.10 and PyTorch 1.11.0+cu113 for development and evaluation.

\textbf{Metrics: }To assess the effectiveness of our federated training approach for diffusion models, we utilize the Fréchet Inception Distance (FID)\cite{fid} to measure the quality and diversity of the generated images. 
We sample 50,000 images to measure the FID.
Additionally, we evaluate the communication efficiency by tracking the total parameters and bytes transferred.
The FID calculates the distance between the mean and covariance of the real and generated image distributions. 
The FID\cite{fid} is defined as:
\begin{equation}
    \text{FID} = \| \mu_r - \mu_g \|^2 + \text{Tr}(\Sigma_r + \Sigma_g - 2(\Sigma_r \Sigma_g)^{\frac{1}{2}}) 
\end{equation}
where $\mu_r$ and $\mu_g$ are the mean vectors, and $\Sigma_r$ and $\Sigma_g$ are the covariance matrices of the real and generated data points, respectively. 
Lower FID  scores indicate that the generated images are more similar to the real images in terms of quality and diversity.
\subsection{Main Results}
In this section, we detail our main findings and evaluate the effectiveness of each of the algorithms described in this work under different conditions related to federated learning.
\subsubsection{RQ1: Federated Averaging for Diffusion Models}
To assess the efficacy of federated averaging (FedAvg) in training diffusion models, we evaluated the DDPM on CIFAR-10, CelebA and LDM-8 on LSUN-Church. We varied several parameters: the total number of clients (5, 10, 15), the proportion of contributing clients per round (0.4 and 0.6), global epochs (2, 4, 8, 16, 20, 30), and local epochs (5, 10, 15). Images generated for different configurations of CelebA have been shown in Figure \ref{fig:federated_averaging_variants} and the results for CIFAR10 and LSUN Church are summarized in Table \ref{tab:Fed-vanilla}. We report selective results which provide a converged view. For 5 clients, both the CIFAR-10 and LSUN-Church have $\text{FIDs}>15$.
For the CIFAR-10 dataset using the DDPM model, the centralized training achieves a Fréchet Inception Distance (FID) score of 5.38, indicating high-quality image generation. Under federated settings, the performance varies with the number of clients and their participation rate. 
Notably, the best FID score of 6.31 is achieved with 15 total clients and 6 contributing clients, utilizing 8 global epochs and 15 local epochs. 
This suggests that increasing the number of clients and local epochs can improve the model's performance, bringing it closer to centralized results. 
In contrast, for the LSUN-Church dataset using the LDM-8 model, centralized training yields a better FID score of 9.07 compared to federated settings. 
The most favorable federated configuration involves 15 total clients, 9 contributing clients, 8 global epochs, and 10 local epochs, resulting in an FID score of 11.79. 
This indicates a substantial degradation in image quality under federated settings for LDMs on high-resolution images.
\subsubsection{RQ2: Image Resolution Impact on Federated Diffusion Models}
\begin{figure}
\centering
\includegraphics[width=0.5\textwidth]{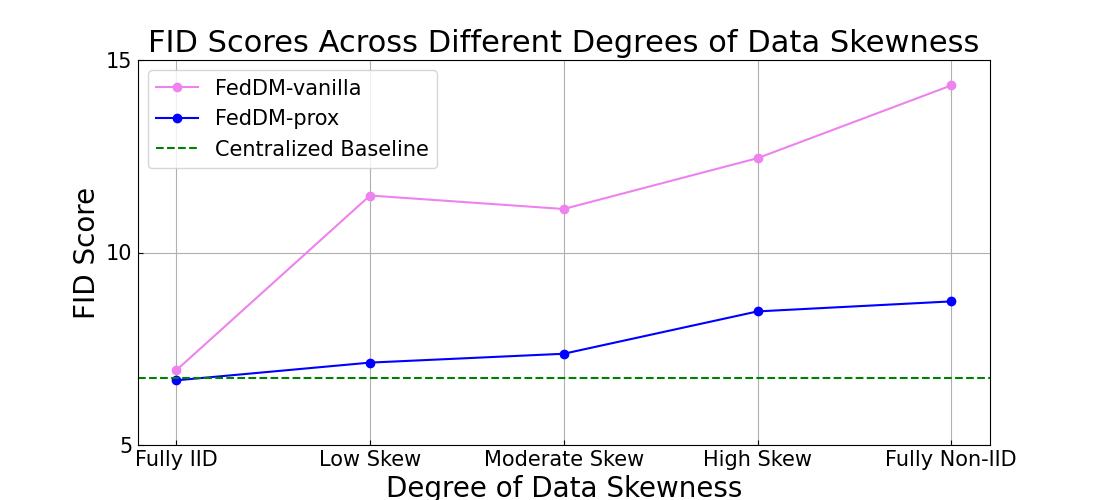}
        \caption{Performance of FedAvg-vanilla and FedAvg-prox for differing degrees of  data skewness for the DDPM model trained on the CelebA dataset}
        \label{fig:fid_scores}
\end{figure}
We investigated the scalability of federated diffusion models by evaluating their performance across various image resolutions. The evaluations were conducted using the DDPM model for datasets such as Fashion MNIST (28x28), CIFAR-10 (32x32), and CelebA (64x64), while the LDM-8 model was assessed on LSUN Church (256x256). These assessments were performed with datasets distributed across 10 clients, of which 6 actively contributed to the training. The models were trained using a combination of global and local epochs.
The results, summarized in Table \ref{tab:scaling}, indicate that federated DDPMs maintain relatively consistent performance across different resolutions. However, there is a noticeable degradation in the generation quality at higher resolutions. Specifically, for the LSUN Church dataset with a resolution of 256x256, the Federated LDM-8 model exhibited a significant drop in performance compared to its centralized counterpart.
The trend observed across the different datasets suggests that as image resolution increases, maintaining generation quality in a federated learning setting also becomes more difficult. 
For lower resolutions like Fashion MNIST (28x28), CIFAR-10 (32x32) and CelebA(64x64), the Federated DDPM models exhibit a minor increase in FID scores compared to centralized models, indicating a slight drop in generation quality. 
However, as the resolution increases to LSUN Church (256x256), even with LDMs, the gap between federated and centralized models widens significantly. 
\subsubsection{RQ3: IID vs. Non-IID Partitions}
To evaluate the effect of IID and non-IID data partitions on model performance, we evaluated the DDPM on different label skewness degrees on the  CIFAR-10 dataset. 
We partition the CIFAR-10 dataset into \( K \) partitions with varying levels of skewness using a controlled statistical approach. For low, moderate, and high skew settings, we adjust the sample distribution by applying a skew factor \( S = 2^{(\text{skew\_level} - 1)} \), where a higher skew level results in one partition receiving a significantly larger portion of the data. In contrast, the remaining partitions get smaller portions. Specifically, for each label, (K-1) partitions receive \( \left\lfloor \frac{N_l}{S + K-1} \right\rfloor \) samples, while the tenth partition receives the remaining samples. This creates increasing levels of label imbalance. The completely non-IID setting assigns all samples of each label to a single partition, maximizing skewness. In contrast, the completely IID setting ensures equal label representation across all partitions.

The results are illustrated in Figure \ref{fig:fid_scores}. 
Our analysis indicates a significant FID degradation in the quality of generated images under non-IID partitions when using FedDM-vanilla.
In contrast, we are adding the proximal term in the local objective( as in  FedDM-prox) remarkably improved generation quality with considerable reductions in FID scores.
\subsubsection{RQ4: Quantization of Model Updates}
Quantization of model updates is a very well-used strategy for achieving communication efficiency. 
We evaluated the effect of quantizing model updates for  8-bit and 16-bit quantization for model weights using the DDPM and LDM-8 models. 
The results for the CIFAR-10 and LSUN Church datasets are illustrated in Table \ref{tab:quant}. CelebA generations for FedDM-vanilla with 16 bits for weights and FedDM-quant for 16 bits for weights are shown in Figure \ref{fig:federated_averaging_variants}(d). We use full precision for activations.
Our evaluations indicate that simply quantizing model updates for Fed-vanilla is inefficient; as the bitwidth is reduced, a noticeable decline in quality is observed. 
However, quantizing and calibrating each layer using Fed-Quant yields results very similar to centralized training, even with 8-bit weights. Sampling using Fed-quant showcases very good performance.
Higher levels of quantization for FedDM-quant, such as 8-bit, demonstrate a significant trade-off: while the number of bytes transferred is reduced fourfold, model performance is degraded, but not as much as quantizing without calibrating, as undertaken by FedDM-vanilla.

\section{Conclusions}
In this work, we present novel federated training algorithms in the FedDM framework. This enables multiple organizations
to pool their data together and train data-demanding diffusion models. We showcase the applicability of FL for diffusion models across multiple dimensions - image resolution, degree of label skewness, and quantized model updates. 
 The performance of our algorithms presents a viable FL approach for diffusion models. While introducing a slight trade-off in generation quality, they enable the significant benefits of privacy-preserving data pooling and robust data-sharing capabilities.
Advancing the field of federated diffusion models necessitates exploring several key areas. First, while scalable privacy attacks have not been implemented yet, performing a privacy analysis of federated learning diffusion models upon their emergence will be crucial. Exploring different types of diffusion models, including those for audio and video, and the applicability to conditional diffusion models are promising directions.

% \bibliographystyle{unsrt}  
% \bibliography{references}

\end{document}